\documentclass[letterpaper]{article} 
\usepackage{aaai25}  
\usepackage{times}  
\usepackage{helvet}  
\usepackage{courier}  
\usepackage[hyphens]{url}  
\usepackage{graphicx} 
\usepackage{natbib}  
\usepackage{caption} 
\usepackage{algorithm}
\usepackage{algorithmic}
\usepackage{amsmath}
\usepackage{amssymb}
\usepackage{adjustbox}
\usepackage{multirow}
\usepackage{booktabs}
\usepackage{dsfont}
\usepackage{subcaption}
\usepackage{xcolor}
\usepackage{hyperref}
\hypersetup{hidelinks,
	colorlinks=true,
        citecolor=black,
        linkcolor=red,
	pdfstartview=Fit,
	breaklinks=true}

\usepackage{newfloat}
\usepackage{listings}
\DeclareCaptionStyle{ruled}{labelfont=normalfont,labelsep=colon,strut=off} 
\floatstyle{ruled}
\newfloat{listing}{tb}{lst}{}
\floatname{listing}{Listing}
\title{Exploring Semantic Consistency and Style Diversity for\\ Domain Generalized Semantic Segmentation}
\author{
    Hongwei Niu\textsuperscript{\rm 1,\rm 2}\equalcontrib, Linhuang Xie\textsuperscript{\rm 1}\equalcontrib, Jianghang Lin\textsuperscript{\rm 1}\equalcontrib,
    Shengchuan Zhang\textsuperscript{\rm 1}\thanks{Corresponding Author.}
}
\affiliations{
    \textsuperscript{\rm 1}Key Laboratory of Multimedia Trusted Perception and Efficient Computing,\\
    Ministry of Education of China, Xiamen University, 361005, P.R. China\\
    \textsuperscript{\rm 2}Institute of Artificial Intelligence, Xiamen University, Fujian, China\\

    \{niuhw649, hunterjlin007\}@stu.xmu.edu.cn\\
    linhuangxie@gmail.com, zsc\_2016@xmu.edu.cn\\
}

\usepackage{bibentry}

\begin{document}
\maketitle
\begin{abstract}
Domain Generalized Semantic Segmentation (DGSS) seeks to utilize source domain data exclusively to enhance the generalization of semantic segmentation across unknown target domains.
Prevailing studies predominantly concentrate on feature normalization and domain randomization, these approaches exhibit significant limitations.
Feature normalization-based methods tend to confuse semantic features in the process of constraining the feature space distribution, resulting in classification misjudgment.
Domain randomization-based methods frequently incorporate domain-irrelevant noise due to the uncontrollability of style transformations, resulting in segmentation ambiguity.
To address these challenges, we introduce a novel framework, named \textbf{SCSD} for \textbf{S}emantic \textbf{C}onsistency prediction and \textbf{S}tyle \textbf{D}iversity generalization.
It comprises three pivotal components:
Firstly, a Semantic Query Booster is designed to enhance the semantic awareness and discrimination capabilities of object queries in the mask decoder, enabling cross-domain semantic consistency prediction.
Secondly, we develop a Text-Driven Style Transform module that utilizes domain difference text embeddings to controllably guide the style transformation of image features, thereby increasing inter-domain style diversity.
Lastly, to prevent the collapse of similar domain feature spaces, we introduce a Style Synergy Optimization mechanism that fortifies the separation of inter-domain features and the aggregation of intra-domain features by synergistically weighting style contrastive loss and style aggregation loss.
Extensive experiments demonstrate that the proposed SCSD significantly outperforms existing state-of-the-art methods. 
Notably, SCSD trained on GTAV achieved an average of 49.11 mIoU on the four unseen domain datasets, surpassing the previous state-of-the-art method by +4.08 mIoU.
Code is available at \url{https://github.com/nhw649/SCSD}.
\end{abstract}
\section{Introduction}
\begin{figure}[t]
\centering
\includegraphics[width=0.48\textwidth]{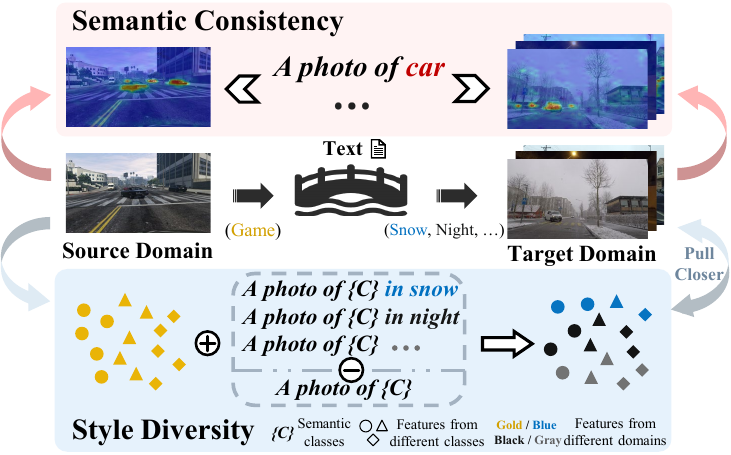}
\caption{Illustration of our motivations. \textbf{Semantic Consistency}: Similarity map between domain-irrelevant general text embeddings and image features in different domains. Cross-domain consistent prediction can be achieved through general text prompts. \textbf{Style Diversity}: Simplified version of t-SNE visualization of image features. Domain text difference embeddings are used as style difference features guide the style transformation of image features from source domain to target domain.}
\label{mot}
\end{figure}
\begin{figure}[t]
\centering
\includegraphics[width=0.48\textwidth]{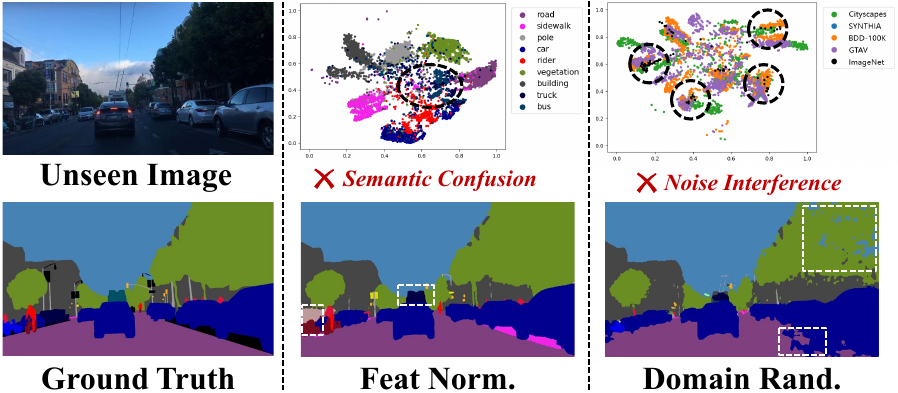} 
\caption{Column 1: Image and ground truth (GT). Columns 2-3: Feature normalization-based and domain randomization-based methods. For both methods, Row 1 shows the t-SNE visualization of feature maps, and Row 2 shows the segmentation result. \textbf{Better view in zoom.}}
\label{prob}
\end{figure}

Semantic segmentation, a core task in computer vision, involves assigning a semantic class label to each pixel in an image.
Traditionally, models~\cite{strudel2021segmenter, li2022deep_seg, mi2022active, hu2023yoso, yue2024adaptive} for object detection and image segmentation are trained and evaluated under the assumption that datasets are independent and identically distributed.
However, this assumption often fails to hold in real-world scenarios due to variations in lighting, weather conditions, and geographical differences. 
This challenge has spurred significant research in Domain Adaptive Semantic Segmentation (DASS)~\cite{hoyer2022daformer_dass, xia2023cmda_dass, cheng2023adpl_dass, choe2024open_set_dass}, which seeks to minimize the distribution discrepancies between source and target domains by aligning their data distributions.
Unlike DASS, which requires access to target domain data during training, an often impractical requirement, Domain Generalized Semantic Segmentation (DGSS) offers a promising alternative.
DGSS focuses on learning only from source domain data to enhance the model's ability to generalize to unknown target domains, thus better meeting the challenges posed by real-world applications.
Existing DGSS methods are generally divided into two categories: feature normalization and domain randomization.
Feature normalization-based methods include normalization~\cite{pan2018IBN, peng2022san-saw, huang2023spc-net} and whitening~\cite{pan2018SW, huang2019Iternorm, choi2021ISW}, aim to constrain the distribution of different features to the same space or remove the domain-specific style features, thereby facilitating the learning of domain-invariant semantic content representations.
However, as illustrated in the second column of  Fig.~\ref{prob}, this often confuse intra-domain semantic features of different categories, resulting in classification misjudgment.
Moreover, due to the non-orthogonality of content and style, removing style can also lead to the loss of semantic content.
Domain Randomization-based methods~\cite{lee2022wildnet, 2023PASTA, fahes2024FAMix} seek to transform the source domain style into multiple other domains to increase style diversity.
However, these methods rely on artificially created auxiliary domains (\textit{e.g.}, ImageNet). As depicted in the third column of Fig.\,\ref{prob}, this can introduce domain-irrelevant noise, potentially compromising the domain-invariant representations and resulting in segmentation ambiguities.
The emergence of Visual-Language Models (VLMs), such as CLIP~\cite{radford2021_CLIP}, represents a significant development in multimodal learning.
These models map images and text into a unified representation space, achieving modality alignment by comparing the similarities and differences between various image-text pairs.
Such capabilities offer new perspectives for the Domain Generalized Semantic Segmentation (DGSS) field.
By harnessing the inherent alignment properties of VLMs, the text modality can effectively act as a conduit to bridge the domain gap between source and target domains.
Specifically, we explore two properties of the text modality in VLMs for enhancing DGSS: \textbf{1) Semantic Consistency}: Despite the significant variability in data distribution across domains, semantic categories remain consistent.
For example, ``a car in game'' and ``a car in snow'' are classified under the same semantic category.
As illustrated in the pink section of Fig.\,\ref{mot}, the alignment capabilities of VLMs between image and text modalities enable a general text prompt (\textit{e.g}., ``a photo of car") to consistently correspond with the target object (car) across various domains (\textit{e.g}., game, snow, night, etc.).
Thus, the text modality can achieve semantically consistent predictions across domains while preserving semantic content.
\textbf{2) Style Diversity}: By aligning images and texts within a unified representation space, VLMs also learn the differences between various image-text pairs.
As shown in the blue section of Fig.\,\ref{mot}, a simplified version of t-SNE visualization (see Experiments for more details) of the original image features and the image features weighted by text difference embeddings demonstrates that differences in text prompt embeddings across domains can be leveraged to diversify the style of image features.
Based on these observations, we introduced a novel framework named \textbf{SCSD}, which is designed to explore \textbf{S}emantic \textbf{C}onsistency and \textbf{S}tyle \textbf{D}iversity for DGSS.
It comprises three carefully designed innovative components that fully exploit the potential of the text modality.
Firstly, we propose Semantic Query Booster (SQB), which leverages semantic consistency between image and text modalities to establish cross-modal semantic associations and aggregate relevant semantic features.
By enhancing the semantic discernment of object queries within the mask decoder, SQB facilitates robust predictions of semantic consistency across different domains.
Secondly, we introduce a Text-Driven Style Transformation (TDST) module that mines the style diversity of the text modality.
By utilizing the difference between text embedding vectors from specific domain prompts and general domain prompts as domain difference embeddings and mapping them to be style difference features, this module can controllably guide the transformation of the low-frequency amplitude spectrum of image features, thereby achieving cross-domain style transformation and enhancing inter-domain style diversity.
Lastly, to prevent the collapse of similar domain feature spaces, we introduce a Style Synergy Optimization mechanism that fortifies the separation of inter-domain features and the aggregation of intra-domain features by synergistically weighting style contrastive loss and style aggregation loss.
We conduct comprehensive experiments in both single-source and multi-source settings to demonstrate that SCSD exhibits superior generalization compared to existing DGSS methods.
Notably, SCSD outperforms the state-of-the-art methods by up to +4.87 and +2.92 mIoU on unseen Mapillary and ACDC domains, respectively.
\section{Related Work}
\subsection{Domain Generalized Semantic Segmentation}
To address the challenges of domain shift and the absence of target domain data, several domain generalization (DG) methods~\cite{Shu_2021_CVPR_meta_learning, kang2022style_data_aug, Chen_2023_ICCV_invariant, dayal2024madg_invariant} have been extensively studied.
They aim to train the model using data from a single or multiple related but different source domains so that it can be generalized to any out-of-distribution target domain.
Domain Generalization Semantic Segmentation (DGSS) extends the domain generalization to a more challenging fine-grained segmentation task.
Existing methods primarily focus on normalization~\cite{pan2018IBN, peng2022san-saw, huang2023spc-net, ahn2024BlindNet}, whitening~\cite{pan2018SW, huang2019Iternorm, choi2021ISW}, and domain randomization~\cite{yue2019DRPC, peng2021GTR, zhong2022advstyle, wu2022siamdoge, lee2022wildnet, kim2023TLDR, 2023PASTA, hoyer2023domain, fahes2024FAMix, DIDEX, jia2025dginstyle}.
For example, SPC-Net~\cite{huang2023spc-net} introduce style projection and semantic clustering to achieve better representation.
TLDR~\cite{kim2023TLDR} learns texture and shape features to mitigate the domain gap problem.
BlindNet~\cite{ahn2024BlindNet} decouples content and style through covariance alignment and semantic consistency contrastive learning.
DIDEX~\cite{DIDEX} employs a diffusion model to generate pseudo target domains with diverse text prompts.
CMFormer~\cite{bi2024CMFormer} introduces a content-enhanced mask attention mechanism and multi-resolution feature fusion strategy to improve the model's adaptability to style variations.
\subsection{Vision-Language Models}
Visual-Language Models (VLMs) such as CLIP~\cite{radford2021_CLIP} leverage web-scale image-text pairs to align visual and textual modalities through contrastive learning, thereby demonstrating strong zero-shot generalization capabilities~\cite{qu2023sg, gong2024cross, zhang2024region_clues_clip, lin2024weakly, qu2024goi} and robustness to natural distribution shifts~\cite{ming2022delving, tu2024closer} in various downstream tasks.
For example, FC-CLIP~\cite{yu2024fc_clip} leverages the powerful open-vocabulary classification capabilities of the frozen CNN-based CLIP, combined with a mask generator, to significantly enhance the performance of open-vocabulary panoptic segmentation.
CLIP-RC~\cite{zhang2024region_clues_clip} explores the use of regional cues to translate image-level knowledge into pixel-level understanding, enabling zero-shot semantic segmentation.
Our work further explores the potential of text modality for DGSS.
\section{Method}
\begin{figure*}[!htp]
\includegraphics[width=\textwidth]{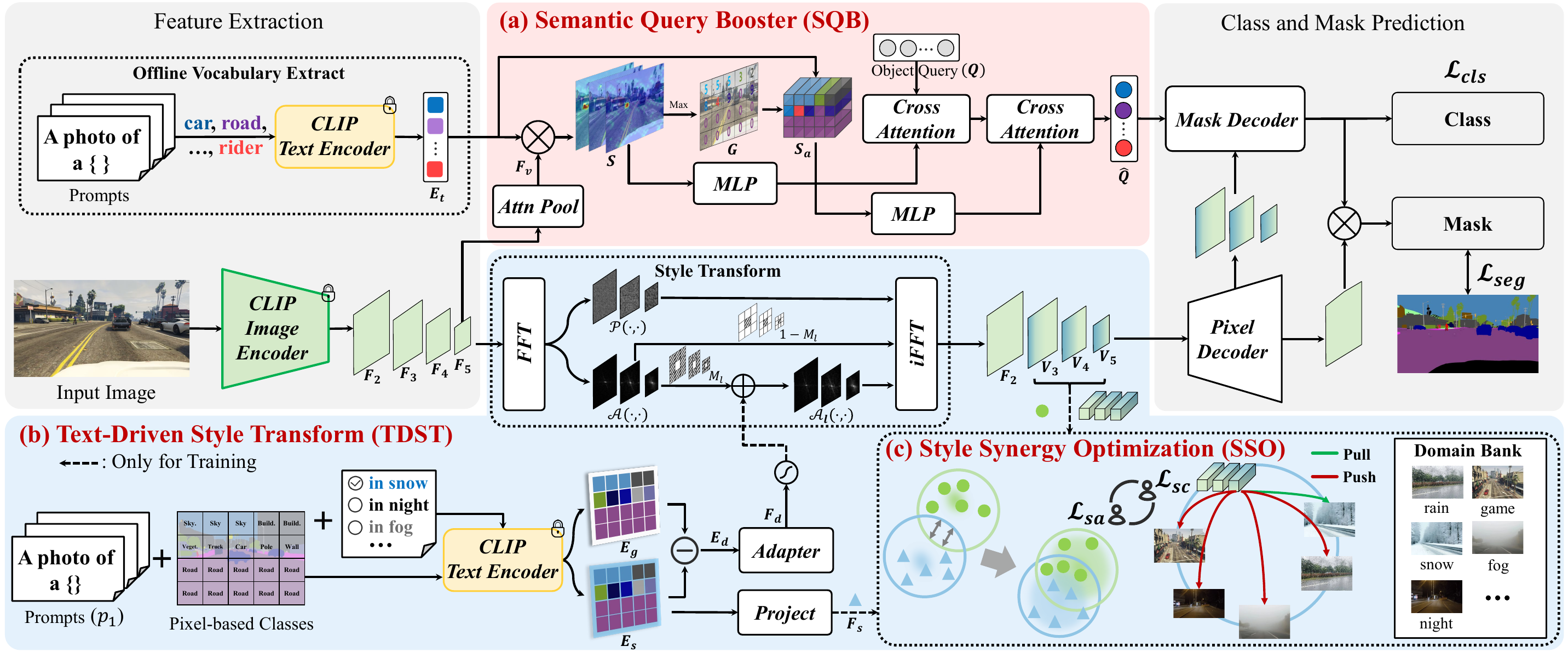}
\caption{Overview of our proposed SCSD. The main contribution are: (a) Semantic Query Booster enhances object queries for cross-domain semantic consistency prediction. (b) Text-Driven Style Transform leverages the diversity of the text modality to facilitate the style transformation of image features. (c) Style Synergy Optimization fortifies the separation of inter-domain features and the aggregation of intra-domain features through the collaboration of two loss functions (\textit{i.e.}, $\mathcal{L}_{sc}$ and $\mathcal{L}_{sa}$).}
\label{framework}
\end{figure*}
\subsection{Overview}
The overall framework of our proposed SCSD is depicted in Fig.\,\ref{framework}, consisting of three key components: Semantic Query Booster (SQB), Text-Driven Style Transform (TDST), and Style Synergy Optimization (SSO).
Initially, for a given input image $I$, multi-scale image features \(F_{i}\in\mathbb{R}^{D\times \frac{H}{2^i}\times \frac{W}{2^i}}\) for \(i \in \{2, 3, 4, 5\}\) are extracted using the CLIP image encoder.
Prompt text embeddings \(E_t\in\mathbb{R}^{C\times D}\), corresponding to $C$ categories from the training dataset, are derived from the CLIP text encoder.
Then, SQB leverages the semantic consistency between image and text modalities to boost object queries \(Q\) into semantic queries \(\hat{Q}\).
TDST mines text style diversity and achieves latter three scale image features style transformation through style difference features obtained by domain difference embeddings mapping.
Finally, in the mask decoder, the semantic queries undergo multiple interactions with the multi-scale image features and continuous refinement, resulting in class and mask predictions.
The overall loss function is defined as:
\begin{equation}
\mathcal{L}=\mathcal{L}_{cls}+\mathcal{L}_{seg}+\mathcal{L}_{s},
\end{equation}
where \(\mathcal{L}_{cls}\) is the cross-entropy loss and \(\mathcal{L}_{seg}\) includes the binary cross-entropy loss and dice loss.
\(\mathcal{L}_{s}\) includes style contrastive loss $\mathcal{L}_{sc}$ and style aggregation loss $\mathcal{L}_{sa}$ in SSO.
\subsection{Semantic Query Booster}
To achieve a consistent understanding of cross-domain semantic knowledge, we propose the SQB, as illustrated in the pink section of Fig.~\ref{framework}.
It leverages domain-irrelevant general text embeddings to establish semantic associations with image features across any domains.
This enables the object queries \(Q\) in the mask decoder to capture the concept of cross-domain semantic consistency and integrate the rich discriminative features learned from various domains, thereby enhancing these semantic awareness and discrimination capabilities of object queries \(Q\).
Specifically, the last layer of image features \(F_5\in\mathbb{R}^{D\times H'\times W'}\) is processed through attention pooling to obtain dense visual features \(F_v\in\mathbb{R}^{H' \times W'\times D}\), where \(H'\) and \(W'\) represent \(\frac{H}{32}\) and \(\frac{W}{32}\), respectively.
This allows for the aggregation of global context information while maintaining the feature scale, thereby providing fine-grained feature representations that more effectively support dense visual tasks.
The semantic similarity map \(S\in\mathbb{R}^{H'\times W'\times C}\) between the dense visual features \(F_v\) and the text embeddings \(E_t\in\mathbb{R}^{C\times D}\) is computed as: $S = F_v\cdot E_t^T$.
It encodes the correlation between each pixel and all categories, which enhance the cross-domain semantic awareness capabilities of the object queries \(Q\), allowing for the correction of semantic bias within the model across different domains and achieving cross-domain semantic consistency prediction.
Subsequently, the most relevant category index \(G\in\mathbb{R}^{H'\times W'}\) for each pixel is determined by performing a maximum operation along the category dimension of the semantic similarity map \(S\).
The corresponding category embedding is retrieved from the text embeddings \(E_t\) using the category index \(G\), and a semantic aggregation map \(S_a\in\mathbb{R}^{H'\times W'\times D}\) tailored for pixel-level classification is constructed:
\begin{equation}
    G=\arg\max_{C}S, 
\end{equation}
\begin{equation}
    S_a = E_t[G],
\end{equation}
where \(E_t[G]\) denotes the selection of the indices in \(G\) along the first dimension of \(E_t\).
The semantic aggregation map \(S_a\) aggregates text semantic embeddings at the pixel level to model a fine-grained semantic space, thereby facilitating the object queries \(Q\) to capture rich semantic information and improving semantic discrimination capabilities.
Both the semantic similarity map \(S\) and the semantic aggregation map \(S_a\) are mapped to the same channel dimensions as the object queries \(Q\) through a MLP respectively.
Finally, a set of learnable object queries \(Q\) interact sequentially with the semantic similarity map \(S\) and the semantic aggregation map \(S_a\) through cross-attention mechanisms to generate the semantic queries \(\hat{Q}\in\mathbb{R}^{N\times D}\) that possess both semantic awareness and discrimination capabilities for cross-domain consistency prediction.
\subsection{Text-Driven Style Transform}
The visual-language models are able to map language and visual information to each other in a unified representation space due to its inherent multimodal alignment capabilities.
Building on it, we leverage the difference features between domain text embeddings to construct style features tailored to each domain.
These style features are used to guide the transformation of image feature styles, ensuring that only the style is altered without affecting the content.
\\
Initially, we introduce a triplet domain prompts set \( P = \{ p_1, p_2, p_3 \} \), which consists of the following:
\begin{itemize}
\item \textbf{General domain prompts} \( p_1 \), such as ``a photo of \{Class\}",``This is a photo of a \{Class\}", etc.
\item \textbf{Conditional domain prompts} \( p_2 \), such as ``\(\phi\)", ``in snow", ``in night", ``in fog", etc, where \(\phi\) represents a null character used to maintain the style of the source domain.
\item \textbf{Specific domain prompts} \( p_3 \), such as ``a photo of \{Class\} \{Domain\}", ``This is a photo of a \{Class\} \{Domain\}", etc.
The \{Domain\} is derived from the conditional domain prompts.
\end{itemize}
To transfer the semantic knowledge of CLIP to pixel-level tasks, domain prompts based on each pixel category are constructed.
Specifically, for each image in the a batch, the specific domain prompts \(p_3\) for the current image are generated by combining the category of each pixel from the ground truth mask with a randomly selected conditional domain prompt \(p_2\).
The general domain prompts \( p_1 \) are constructed directly from the category of each pixel in the ground truth mask.
These two sets of prompts are then separately fed into the text encoder, and the resulting embeddings are averaged along the prompt dimension to obtain the corresponding specific domain embeddings \(E_s\in\mathbb{R}^{H\times W\times C}\) and general domain embeddings \(E_g\in\mathbb{R}^{H\times W\times C}\).
The domain difference embeddings \(E_d\in\mathbb{R}^{H\times W\times C}\) are obtained through element-wise subtraction as follows:
\begin{equation}
E_d = E_s - E_g = \mathcal{E}(p_{3}) - \mathcal{E}(p_{1}),
\end{equation}
where \(\mathcal{E}(\cdot)\) denotes the CLIP text encoder.
To transform the pixel-level text embeddings into the image feature space, we introduce a domain style adapter that maps the domain difference embeddings \(E_d\) to the style difference features \(F_d\in\mathbb{R}^{H\times W\times C}\).
It can be implemented using any unbiased adapter~\cite{chen2024conv_adapter}, or alternatively, with the simplest 1×1 convolution structure.
However, directly integrating the style difference features \(F_d\) with image features may alter the semantic content of the features, leading to prediction bias.
To address this issue, a simple and effective alternative, the Fourier Transform, is worth considering.
Previous research~\cite{yang2020fourier_fda, xu2021fourier_DG} has demonstrated the effectiveness of Fourier Transform-based data augmentation strategies in enhancing model generalization capabilities.
Additionally, since the low-frequency components of the amplitude spectrum encode the overall style and global features of an image, while the phase spectrum encodes the semantic content and spatial distribution.
Therefore, unlike conventional data augmentation, our objective is to use the style difference features as controllable style components within the low-frequency amplitude spectrum of the multi-scale features of images and guide the transformation of feature styles in a controlled manner.
For brevity, we concentrate exclusively on single layer image features \(f\in\mathbb{R}^{D\times H\times W}\) to illustrate our method.
First, Fast Fourier Transform (FFT) is applied to decompose the image features \(f\) into the amplitude spectrum \(\mathcal{A}(\cdot,\cdot)\) and the phase spectrum \(\mathcal{P}(\cdot,\cdot)\), as follows:
\begin{equation}
\begin{aligned}
&\mathcal{A}(u,v)=\sqrt{\mathrm{Re}(\mathcal{F}(u,v))^2+\mathrm{Im}(\mathcal{F}(u,v))^2},\\
&\mathcal{P}(u,v)=\tan^{-1}\left(\frac{\mathrm{Im}(\mathcal{F}(u,v))}{\mathrm{Re}(\mathcal{F}(u,v))}\right),
\end{aligned}
\end{equation}
where \( H \) and \( W \) are the height and width of the image features, \( x \) and \( y \) are the pixel coordinates in the spatial domain, and \( u \) and \( v \) are correspond to the frequency coordinates in the frequency domain.
\(\mathrm{Re}(\cdot)\) and \(\mathrm{Im}(\cdot)\) denote the real and imaginary parts of the Fourier spectrum \(\mathcal{F}(\cdot, \cdot)\).
To enhance the interpretability of the spectrum, the amplitude spectrum is centralized to concentrate the low-frequency components at the center of the spectrum.
Additionally, a low-frequency mask \(M_l\in \{0,1\}\) is constructed to effectively extract the low-frequency components, as defined by:
\begin{equation}
M_l(x, y) = \mathds{1}_{(x, y) \in [\frac H2(1-\alpha) : \frac H2(1+\alpha), \frac W2(1-\alpha) : \frac W2(1+\alpha)]},
\label{eq:low_fre_ratio}
\end{equation}
where \(\alpha\) is the ratio of low-frequency components.
Next, to prevent the introduction of uncontrollable noise resulting from excessive style differences, the style difference features \(F_d\) are numerically constrained by the hyperbolic tangent function.
The activated style difference features are then weighted and summed with the low-frequency components of the amplitude spectrum, and combined with the amplitude spectrum of the high-frequency components to obtain the composite amplitude spectrum \(\mathcal{A}_c(\cdot,\cdot)\), as follows:
\begin{equation}
\mathcal{A}_l(u, v)=M_l(x, y) \cdot \mathcal{A}(u, v)\cdot(1+\beta\cdot \tanh(F_{d})),
\label{eq:intensity}
\end{equation}
\begin{equation}
\mathcal{A}_c(u, v)=(1-M_l(x, y)) \cdot \mathcal{A}(u, v) +  M_l(x, y) \cdot \mathcal{A}_l(u, v),
\notag
\end{equation}
where \(\beta\) is the intensity of style control.
For high-level image features, which contain more abstract characteristics, a stronger style control is applied to alter the overall style.
Conversely, for low-level image features, which  encompass finer details, a weaker style control is employed to better preserve the original content.
Finally, the composite amplitude spectrum \(\mathcal{A}_c(\cdot,\cdot)\) and the original phase spectrum \(\mathcal{P}(\cdot,\cdot)\) are merged and transformed using the inverse Fast Fourier Transform (iFFT) to obtain the style visual features \(V\in\mathbb{R}^{D\times H\times W}\) after style transformation:
\begin{equation}
V(x,y)=\mathcal{F}^{-1}([\mathcal{A}_c(u, v), \mathcal{P}(u,v)]).
\end{equation}
\textbf{Discussion.} We identified several key advantages of TDST as follows:
Unlike methods such as WildNet~\cite{lee2022wildnet}, which rely on wild datasets (\textit{e.g.}, ImageNet) to construct style auxiliary domains, TDST does not require these external datasets.
Moreover, our text-driven style transformation is both precise and controllable, effectively avoiding domain-irrelevant style transformations.
Second, due to the image-text alignment mechanism inherent in Visual-Language Models, TDST leverages rich and controllable text descriptions to drive the model.
This allows it to naturally learn generalized domain information from diverse representations.
Moreover, unlike the two-stage method~\cite{fahes2024FAMix}, our TDST is trained end-to-end and used exclusively during the training phase, significantly improving the inference efficiency of the model.

\begin{table*}[!htp]
\centering
\begin{adjustbox}{max width=0.99\textwidth}
\begin{tabular}{c|c|ccccc|ccccc}
    \toprule
    \multirow{2.5}{*}{Method} & \multirow{2.5}{*}{Backbone} & \multicolumn{5}{c|}{Trained on GTAV (G)} & \multicolumn{5}{c}{Trained on Cityscapes (C)}\\
    \cmidrule(lr){3-12}
    & & $\rightarrow$C & $\rightarrow$B & $\rightarrow$M & $\rightarrow$S & Avg.
    & $\rightarrow$B & $\rightarrow$M & $\rightarrow$G & $\rightarrow$S & Avg.\\
    \midrule
    IBN~\cite{pan2018IBN} & RestNet-50 & 33.85 & 32.30 & 37.75 & 27.90 & 32.95 & 48.56 & 57.04 & 45.06 & 26.14 & 44.20\\
    Iternorm~\cite{huang2019Iternorm} & RestNet-50 & 31.81 & 32.70 & 33.88 & 27.07 & 31.37 & 49.23 & 56.26 & 45.73 & 25.98 & 44.30\\
    ISW~\cite{choi2021ISW} & RestNet-50 & 36.58 & 35.20 & 40.33 & 28.30 & 35.10 & 50.73 & 58.64 & 45.00 & 26.20 & 45.14\\
    SHADE~\cite{zhao2022shade} & RestNet-50 & 44.65 & 39.28 & 43.34 & 28.41 & 38.92 & 50.95 & 60.67 & 48.61 & 27.62 & 46.96\\
    SAN-SAW~\cite{peng2022san-saw} & RestNet-50 & 39.75 & 37.34 & 41.86 & 30.79 & 37.44 & \underline{52.95} & 59.81 & 47.28 & 28.32 & 47.09\\
    WildNet~\cite{lee2022wildnet} & RestNet-50 & 44.62 & 38.42 & 46.09 & 31.34 & 40.12 & 50.94 & 58.79 & 47.01 & 27.95 & 46.17\\
    HRDA (Hoyer et al.~\citeyear{hoyer2023domain}) & RestNet-101 & 39.63 & 38.69 & 42.21 & - & - & - & - & - & - & -\\
    TLDR (Kim et al.~\citeyear{kim2023TLDR}) & RestNet-50 & 46.51 & 42.58 & 46.18 & 36.30 & 42.89 & - & - & - & - & -\\
    DPCL (Yang et al.~\citeyear{yang2023DPCL}) & RestNet-50 & 44.87 & 40.21 & 46.74 & - & - & 52.29 & - & 46.00 & 26.60 & -\\
    HGFormer~\cite{ding2023hgformer} & RestNet-50 & - & - & - & - & - & 51.50 & 61.60 & 50.40 & 30.10 & 48.40\\
    BlindNet~\cite{ahn2024BlindNet} & RestNet-50 & 45.72 & 41.32 & 47.08 & 31.39 & 41.38 & 51.84 & 60.18 & 47.97 & 28.51 & 47.13\\
    FAMix~\cite{fahes2024FAMix} & RestNet-50 & \underline{48.15} & \textbf{45.61} & \underline{52.11} & \underline{34.23} & \underline{45.03} & \textbf{54.07} & 58.72 & 45.12 & 32.67 & 47.65\\
    DGInStyle~\cite{jia2025dginstyle} & RestNet-101 & 46.89 & 42.81 & 50.19 & - & - & - & - & - & -\\
    \midrule
    SCSD (Ours) & RestNet-50 & \textbf{51.72} & \underline{44.67} & \textbf{56.98} & \textbf{43.08} & \textbf{49.11} & 52.25 & \textbf{62.51} & \textbf{51.00} & \textbf{39.77} & \textbf{51.38}\\
    \bottomrule
\end{tabular}
\end{adjustbox}
\caption{\textbf{Single-source setting trained on GTAV.} Mean IoU(\%) comparison of different methods. ``-'' indicates the metric is not reported or the official source code is not available. The \textbf{best} and \underline{second best} results are emphasized.}
\label{tab:g}
\end{table*}
\begin{table}[t]
\centering
\begin{adjustbox}{max width=0.47\textwidth}
\begin{tabular}{c|cccc}
    \toprule
    \multirow{2.5}{*}{Method} & \multicolumn{4}{c}{Trained on Two Datasets (G+S)} \\
    \cmidrule(lr){2-5}
    & $\rightarrow$ C & $\rightarrow$ B & $\rightarrow$ M & Avg.\\
    \midrule
    IBN~\cite{pan2018IBN} & 35.55 & 32.18 & 38.09 & 35.27\\
    ISW~\cite{choi2021ISW} & 37.69 & 34.09 & 38.49 & 36.76\\
    SHADE~\cite{zhao2022shade} & 47.43 & 40.30 & 47.60 & 45.11\\
    Pin the memory~\cite{kim2022pin} & 44.51 & 38.07 & 42.70 & 41.76\\
    AdvStyle~\cite{zhong2022advstyle} & 39.29 & 39.26 & 41.14 & 39.90\\
    TLDR (Kim et al.~\citeyear{kim2023TLDR}) & 48.83 & 42.58 & 47.80 & 46.40\\
    SPC-Net~\cite{huang2023spc-net} & 46.36 & 43.18 & 48.23 & 45.92\\
    FAMix~\cite{fahes2024FAMix} & \underline{49.41} & \textbf{45.51} & \underline{51.61} & \underline{48.84}\\
    \midrule
    SCSD (Ours) & \textbf{52.43} & \underline{45.25} & \textbf{56.58} & \textbf{51.42}\\
    \bottomrule
\end{tabular}
\end{adjustbox}
\caption{\textbf{Multi-source setting trained on GTAV + SYNTHIA.} Mean IoU(\%) comparison of different methods with ResNet-50 backbone.}
\label{tab:g_s}
\end{table}
\begin{table}[t]
\centering
\begin{adjustbox}{max width=0.47\textwidth}
\begin{tabular}{c|ccccc}
    \toprule
    \multirow{2.5}{*}{Method} & \multicolumn{5}{c}{Trained on GTAV (G)} \\
    \cmidrule(lr){2-6}
    & $\rightarrow$ AN & $\rightarrow$ AS & $\rightarrow$ AR & $\rightarrow$ AF & Avg.\\
    \midrule
    ISW~\cite{choi2021ISW} & 6.32 & 29.97 & 33.02 & 32.56 & 25.47\\
    SHADE~\cite{zhao2022shade} & 8.18 & 30.38 & 35.44 & 36.87 & 27.72\\
    WildNet~\cite{lee2022wildnet} & 8.27 & 30.29 & 36.32 & 35.39 & 27.57\\
    SiamDoGe~\cite{wu2022siamdoge} & 10.60 & 30.71 & 35.84 & 36.45 & 28.40\\
    TLDR (Kim et al.~\citeyear{kim2023TLDR}) & 13.13 & 36.02 & \underline{38.89} & \underline{40.58} & 32.16\\
    FAMix~\cite{fahes2024FAMix} & \underline{14.96} & \underline{37.09} & 38.66 & 40.25 & \underline{32.74}\\
    \midrule
    SCSD (Ours) & \textbf{15.06} & \textbf{41.37} & \textbf{42.77} & \textbf{43.43} & \textbf{35.66}\\
    \bottomrule
\end{tabular}
\end{adjustbox}
\caption{\textbf{Adverse Condition setting trained on GTAV.} Mean IoU(\%) comparison of different methods with ResNet-50 backbone. }
\label{tab:acdc}
\end{table}
\begin{table}[t]
\centering
\begin{adjustbox}{max width=0.47\textwidth}
\begin{tabular}{l|ccccc}
    \toprule
    &\multicolumn{5}{c}{Trained on GTAV (G)} \\
    \midrule
    Components & $\rightarrow$ C & $\rightarrow$ B & $\rightarrow$ M & $\rightarrow$ S & Avg.\\
    \midrule
    baseline & 49.16 & 42.19 & 53.73 & 40.54 & 46.41\\
    + SQB & 49.52~\tiny{\color{red} ($\uparrow$ 0.36) } & 43.28~\tiny{\color{red} ($\uparrow$ 1.09) } & 55.17~\tiny{\color{red} ($\uparrow$ 1.44) } & 41.75~\tiny{\color{red} ($\uparrow$ 1.21) } & 47.43~\tiny{\color{red} ($\uparrow$ 1.02) }\\
    ++ TDST & 50.70~\tiny{\color{red} ($\uparrow$ 1.18) } & 44.04~\tiny{\color{red} ($\uparrow$ 0.76) } & 55.71~\tiny{\color{red} ($\uparrow$ 0.54) } & 42.83~\tiny{\color{red} ($\uparrow$ 1.08) } & 48.32~\tiny{\color{red} ($\uparrow$ 0.89) }\\
    +++ SOO & \textbf{51.72}~\tiny{\color{red} ($\uparrow$ 1.02) } & \textbf{44.67}~\tiny{\color{red} ($\uparrow$ 0.63) } & \textbf{56.98}~\tiny{\color{red} ($\uparrow$ 1.27) } & \textbf{43.08}~\tiny{\color{red} ($\uparrow$ 0.25) } & \textbf{49.11}~\tiny{\color{red} ($\uparrow$ 0.79) }\\
    \bottomrule
\end{tabular}
\end{adjustbox}
\caption{Ablation studies on each component of SCSD.}
\label{tab:abl_com}
\end{table}
\begin{table}[t]
\centering
\begin{adjustbox}{max width=0.46\textwidth}
\begin{tabular}{ccc|ccccc}
    \toprule
    \multicolumn{3}{c}{SSO} & \multicolumn{5}{c}{Trained on GTAV (G)} \\
    \midrule
    $\mathcal{L}_{sc}$ & $\mathcal{L}_{sa}$ & $w$ & $\rightarrow$ C & $\rightarrow$ B & $\rightarrow$ M & $\rightarrow$ S & Avg.\\
    \midrule
    - & - & - & 50.70 & 44.04 & 55.71 & 42.83 & 48.32\\
    \checkmark & - & - & 50.47 & 44.25 & 55.94 & 42.87 & 48.38\\
    - & \checkmark & - & 49.86 & 44.38 & 54.56 & 41.72 & 47.63\\
    \checkmark & \checkmark & - & 50.83 & \textbf{45.21} & 56.50 & 42.49 & 48.76\\
    \checkmark & \checkmark & \checkmark & \textbf{51.72} & 44.67 & \textbf{56.98} & \textbf{43.08} & \textbf{49.11} \\
    \bottomrule
\end{tabular}
\end{adjustbox}
\caption{Ablation studies on SSO.}
\label{tab:abl_sso}
\end{table}
\subsection{Style Synergy Optimization}
Domain condition prompts are the key factors in guiding style transformation.
However, a domain collapse issue arises: similar domain condition prompts (such as ``in rain" and ``in hail") are combined and encoded into similar style difference features.
This leads to different style visual features being projected into the same style space distribution, thereby weakening the model's ability to learn and represent a diverse range of domain styles.
To address this problem, we introduced SSO mechanism that synergistically weights the style contrastive loss and style aggregation loss based on their changing trends, further strengthening inter-domain feature separation and intra-domain feature aggregation.
Specifically, a set of learnable domain bank \(D\in\mathbb{R}^{K\times D}\) initialized by domain conditional prompts embeddings are used to store style vectors for each domain, where \(K\) is the number of conditional domain prompts.
The style contrastive loss is designed to pull the style visual features closer to the style vectors of the same domain while pushing them farther from the style vectors of other domains.
Consequently, style contrastive loss is computed between the globally average-pooled style visual features \(\overline{V}\in\mathbb{R}^{1\times D}\) after global average pooling and the domain bank \(D\) as follows:
\begin{equation}
 \!\!\!   \mathcal{L}_{sc} = - \frac{1}{L} \sum_{l=1}^{L} \log 
    (\frac{e^{(\overline{V}_l \cdot D^+) / {\tau}}}{e^{(\overline{V}_l \cdot D^+) / {\tau}} + \sum_{j=1}^{N} e^{(\overline{V}_l \cdot D^-_j) / {\tau}}}),
\end{equation}
where \(L\) is number of feature layers and \(N\) is the number of negative domain samples.
Through this bidirectional optimization, the domain bank can continuously enhance the distinction between style vectors of different domains.
In turn, this process also constrains the style visual features produced by style transformations guided by similar domains, ensuring effective inter-domain feature separation and preventing domain collapse.
To further enhance style alignment, specific domain embeddings are first projected into the image feature space to obtain specific domain style features \(F_s\).
Both the specific domain style features \(F_s\) and the style visual features \(V\) are then normalized to better measure their correlation.
Finally, the relationship is constrained using an L2 loss function:
\begin{equation}
\mathcal{L}_{\mathrm{sa}}=\frac1{LHW}\sum_{l=1}^L\sum_{h=1}^{H}\sum_{w=1}^{W}w_l\cdot\|V^{(l)}_{(h,w)} - F_{s,(h,w)}\|_2^2,
\label{eq:sa}
\end{equation}
where \(w_l\) is weight coefficients that change with the number of layers.
To encourage mutual cooperation between the two losses, our synergy weighting strategy is designed to adaptively adjust the weight of one loss based on changes in the other, as shown below:
\begin{equation}
w = 
\begin{cases} 
w_{init} \times \left(1 - \lambda \times \Delta \mathcal{L}\right), & \text{if } \Delta \mathcal{L} > 0 \\
w_{init} \times \left(1 + \lambda \times |\Delta \mathcal{L}|\right), & \text{otherwise}
\end{cases}
\label{eq:coopera}
\end{equation}
where \(w_{init}\) is the initial weight coefficient, \(\lambda\) is the hyperparameter used for scaling the change.
The idea behind synergy optimization is that when one loss increases, it should take precedence in the optimization process, reducing the contribution of the other loss. As the loss decreases, the weight of the other loss should be increased to promote collaboration between the two losses.
According to Eq.~\ref{eq:coopera}, the weight coefficients for each component are determined, resulting in the overall loss as follows:
\begin{equation}
\mathcal{L}_s=w_{sa}\cdot\mathcal{L}_{sc}+w_{sc}\cdot\mathcal{L}_{sa}.
\end{equation}

\section{Experiments}
\subsection{Settings \& Implementation Details}
\textbf{Datasets.} We conduct the experiments on two synthetic datasets and four real-world datasets to evaluate the generalization capability of our method.
For synthetic datasets, GTAV (G)~\cite{richter2016GTAV} is a game synthetic dataset that contains 24,966 images with a resolution of 1914$\times$1052.
SYNTHIA (S)~\cite{ros2016synthia} is a large-scale synthetic dataset that contains 9,400 images with a resolution of 1280$\times$760.
For real-world datasets, Cityscapes (C)~\cite{cordts2016cityscapes}, BDD-100K (B)~\cite{yu2020bdd100k} and Mapillary (M)~\cite{neuhold2017mapillary} contain 2,975, 7,000 and 18,000 training images and 500, 1,000 and 2,000 validation images, respectively.
ACDC~\cite{sakaridis2021acdc} is a challenging dataset of driving scenes under adverse weather conditions, including: night (AN), snow (AS), rain (AR) and fog (AF) with 106, 100, 100 and 100 validation images respectively.
\\
\textbf{Evaluation Protocols.} Following prior works~\cite{lee2022wildnet, huang2023spc-net, fahes2024FAMix}, the model is trained on the source domain dataset and evaluated on other datasets as the target domain.
Three settings include: 1) G$\rightarrow$\{C, B, M, S\}; 2) C$\rightarrow$\{B, M, G, S\}; and 3) G+S$\rightarrow$\{C, B, M\}.
We use the mIoU (\%) metric for evaluation.
In addition, we report the average mIoU on the target domain datasets.
\\
\textbf{Implementation Details.} 
For fair comparison, we use the same ResNet-50 as the prior methods~\cite{lee2022wildnet, yang2023DPCL, fahes2024FAMix} as the backbone of the segmentation model, which is initialized with CLIP pre-trained weights and perform architecture surgery following~\cite{li2023CLIP_Surgery}.
For the pixel decoder and mask decoder, we follow the default settings in~\cite{ding2023hgformer, bi2024CMFormer}.
The ratio of low-frequency components \(\alpha\) is set to 0.15 in Eq.~\ref{eq:low_fre_ratio}.
The intensity of style control \(\beta\) is set to [1.0, 2.0, 4.0] in Eq.~\ref{eq:intensity}.
The weight coefficients \(w_l\) is set to [0.2, 0.5, 1.0] in Eq.~\ref{eq:sa}.
The initial weight coefficient \(w_{init}\) is set to 1, and the hyperparameter \(\lambda\) in Eq.~\ref{eq:coopera} is set to 0.3.
The model is trained for 90K iterations with a batch size of 2.
All experiments are conducted on a single machine equipped with two NVIDIA 3090 GPUs, each with 24 GB of memory.
For more implementation details, reference to the \textit{Appendix}.
\subsection{Comparison with the state-of-the-art methods}
\textbf{Single-source setting.} Tab.~\ref{tab:g} shows the generalization performance of the models under G$\rightarrow$\{C, B, M, S\} setting and C$\rightarrow$\{B, M, G, S\} setting.
For G$\rightarrow$\{C, B, M, S\} setting, our method achieves an average of 49.11 mIoU on four target domain datasets, yielding an improvement of +4.08 mIoU over the state-of-the-art DGSS methods.
For C$\rightarrow$\{B, M, G, S\} setting, compared to the state-of-the-art DGSS methods, SCSD achieves improvements in mIoU by +0.91, +0.6, and +7.1 on the Mapillary, GTAV, and SYNTHIA datasets, respectively.
These experimental results show that SCSD demonstrates good generalization capabilities, whether trained on real datasets or synthetic datasets.
%
\\
\textbf{Multi-source setting.} Tab.~\ref{tab:g_s} shows the generalization performance of the models under G+S$\rightarrow$\{C, B, M\} setting.
Even when trained exclusively on synthetic datasetss, our method demonstrates a significantly superior generalization capabilities, consistently outperforming other methods across all real-world datasets, with an average improvement of +2.58 mIoU compared to the state-of-the-art methods.
\\
\textbf{Adverse condition settings.} In Tab.~\ref{tab:acdc}, we further validate the generalization performance of SCSD on the challenging adverse condition dataset~\cite{sakaridis2021acdc}.
Specifically, our method improves mIoU by +0.10, +4.28, +4.11, and +3.18 on night (AN), snow (AS), rain (AR) and fog (AF), respectively.
This further demonstrates the strong generalization capability of SCSD.
\subsection{Ablation Studies}
\textbf{Key Components.}
We study the individual contribution of each proposed component to the overall performance.
Tab.~\ref{tab:abl_com} shows the average mIoU improvement on SCSD as we progressively integrate the three key components.
Specifically, the model with SQB (row 2) shows better results compared to the baseline model (row 1), with an average gain of +1.02 mIoU.
This indicates that semantic consistency is crucial for cross-domain robust prediction.
The TDST (row3) introduced only during the training phase improves the performance by +0.89 mIoU on average, which demonstrates the importance of style diversity for the model to generalize to the wide range of domains.
Finally, SSO (row4) is added to further improve the performance by +0.79 mIoU on average.
\\
\textbf{Style Synergy Optimization.}
In Tab.~\ref{tab:abl_sso}, we find that when only the style aggregation loss is applied (row 3), it merely focuses on aggregating style visual features and specific domain style features. As a result, it fails to distinguish between the styles of similar domains, which exacerbates the negative impact of domain collapse and leads to performance degradation.
When the style contrastive loss and synergy weighting strategy are gradually introduced (row 4 and 5), the two losses work together to strengthen the separation of inter-domain features and the aggregation of intra-domain features, and the performance is further improved.
\begin{figure}[t]
\centering
\includegraphics[width=0.48\textwidth]{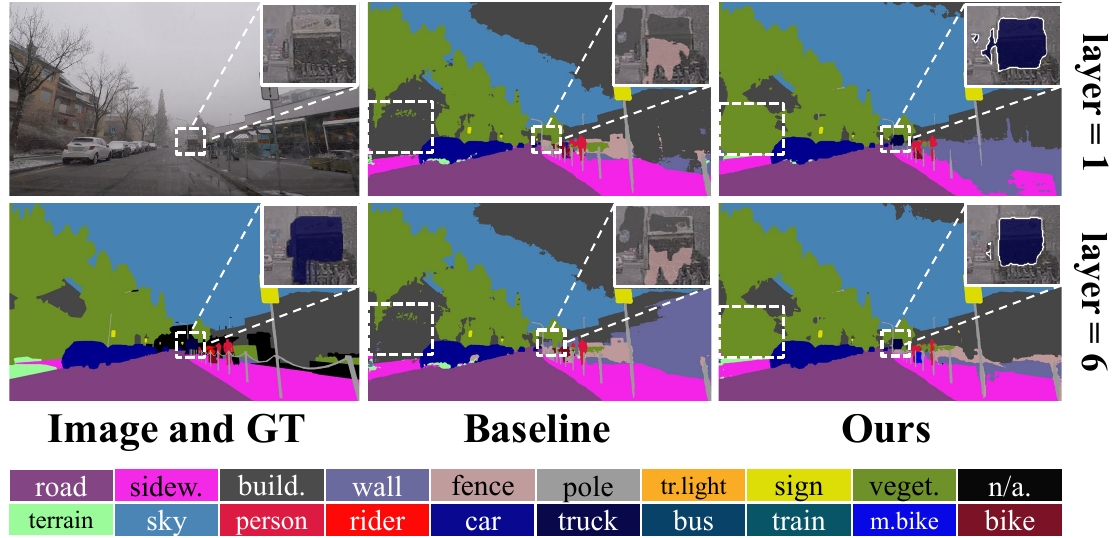}
\caption{Column 1: Image and ground truth (GT). Columns 2-3: Prediction masks output by baseline without SQB and our SCSD at the first and sixth layers of the mask decoder.}\label{SQB_visual}
\end{figure}
\begin{figure}[t]
    \centering
    \begin{subfigure}[b]{0.23\textwidth}
        \centering
        \includegraphics[width=\textwidth]{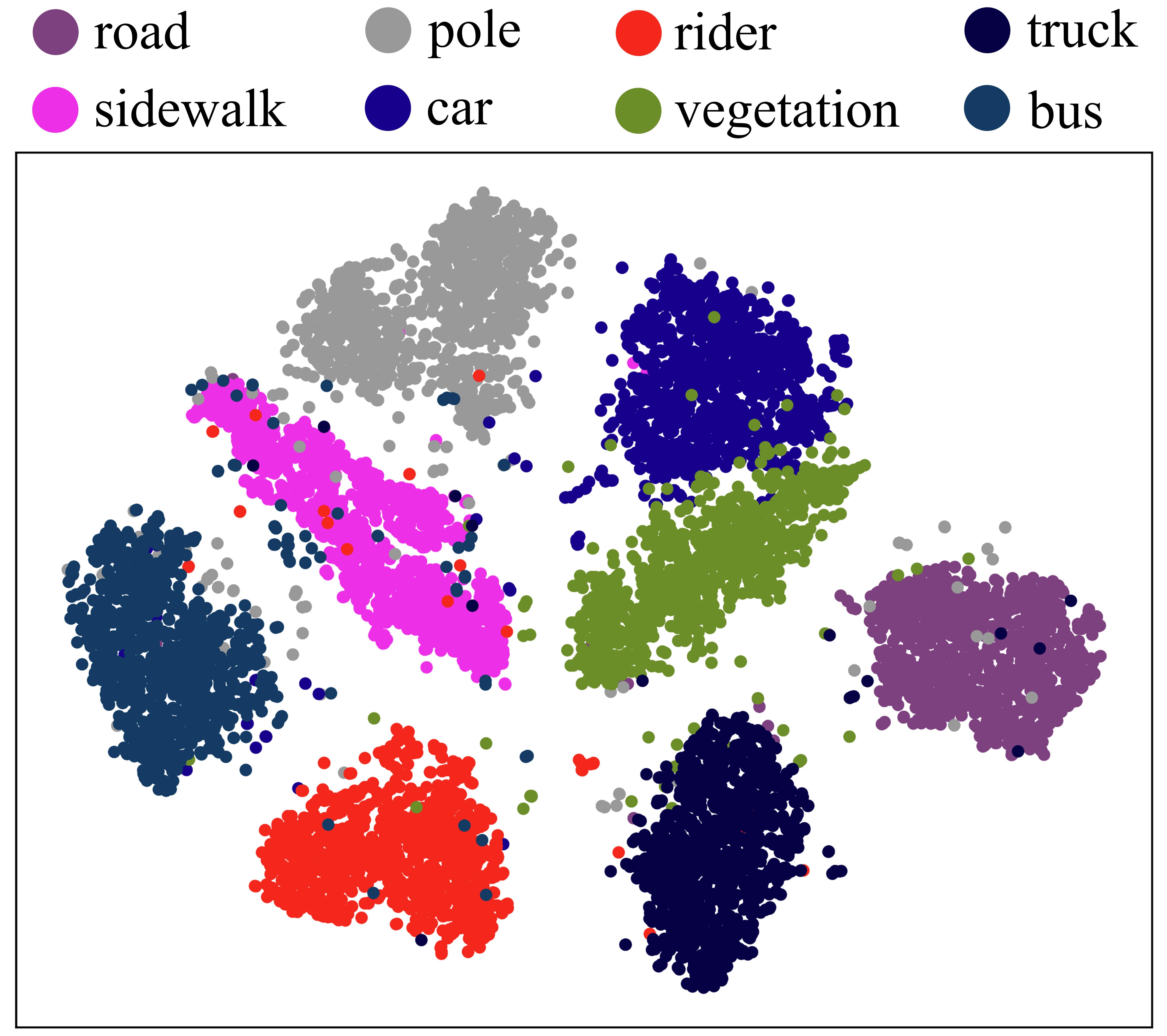}
        \caption{Class distribution}
        \label{class_dis}
    \end{subfigure}
    \hfill
    \begin{subfigure}[b]{0.23\textwidth}
        \centering
        \includegraphics[width=\textwidth]{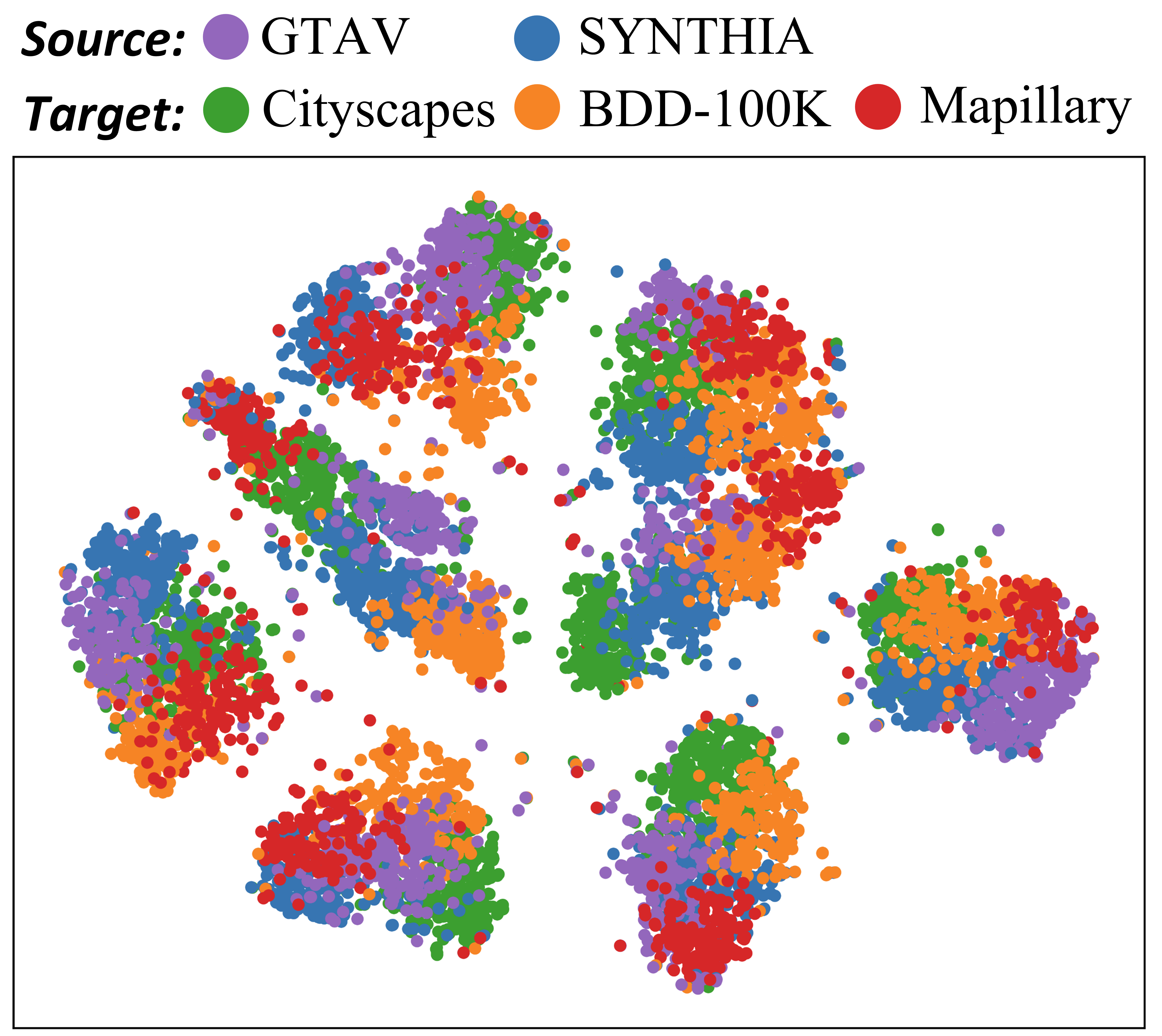}
        \caption{Domain distribution}
        \label{domain_dis}
    \end{subfigure}
    \caption{t-SNE visualization of image features. Colors denote different categories in (a) and different domains in (b).}
    \label{tsne_visual}
\end{figure}
\subsection{Qualitative Analysis.}
\textbf{Benefits from Semantic Query Booster.} In Fig.~\ref{SQB_visual}, we visualize the mask predictions of the baseline and SCSD outputs at different layers of the mask decoder.
In the second column of Fig.~\ref{SQB_visual}, we observe that when dealing with challenging domains, such as snow, whether it's the initial mask output by the first layer of the mask decoder or the refined mask from the sixth layer, the baseline consistently confuses the "vegetation" and "building" categories (refer to the white box in the lower-left corner).
Additionally, the ``truck" consistently lacks a clear mask prediction (refer to the upper-right white box).
In the third column of Fig.~\ref{SQB_visual}, it is incredible that after being equipped with SQB, SCSD generates the accurate initial masks even in the first layer, allowing for further refinement in subsequent layers.
We consider this improvement is due to the object queries learning semantic associations between the visual and textual modalities, enabling cross-domain consistency in predictions, even in unseen extreme domains.
\\
\textbf{Distribution analysis.} We conducted t-SNE visualizations of the image features used for classification to analyze the effectiveness of our SCSD.
Fig.~\ref{class_dis} demonstrates that our method effectively enables the model to distinguish between different categories, significantly reducing semantic confusion caused by the convergence of features from different categories within the same domain.
Furthermore, as shown in Figure ~\ref{domain_dis}, the model successfully clusters samples from different domains according to their respective categories.
This success can be attributed to our proposed TDST module, which allows the model to learn domain-specific style knowledge through domain text difference features, even without the use of additional domain data.
See the \textit{Appendix} for more qualitative analysis.
\section{Conclusion}
In this paper, we propose a novel Semantic Consistency and Style Diversity (SCSD) framework to addresses the limitations of existing methods in Domain Generalized Semantic Segmentation (DGSS).
To effectively leverage semantic consistency between image and text modalities, and the style diversity of the text modality, we introduce three innovative components: the Semantic Query Booster (SQB), the Text-Driven Style Transform (TDST) module, and the Style Synergy Optimization (SSO) mechanism.
Extensive experiments on multiple benchmarks datasets demonstrated the significant performance improvements achieved by SCSD, setting new state-of-the-art results for DGSS.
\section{Acknowledgments}
This work was supported by National Science and Technology Major Project (No.2022ZD0118201), the National Science Fund for Distinguished Young Scholars (No.62025603), the National Natural Science Foundation of China (No.U21B2037, No.U22B2051, No.U23A20383, No.U21A20472, No.62176222, No.62176223, No.62176226, No.62072386, No.62072387, No.62072389, No.62002305 and No.62272401), and the Natural Science Foundation of Fujian Province of China (No.2021J06003, No.2022J06001).
\bibliography{aaai25}

\clearpage
\appendix
\section{Overview}
In this supplementary material, we provide additional experimental results and analyses.
\begin{itemize}
    \item More Implementation Details.
    \item More Quantitative Analysis.
    \item More Qualitative Analyses.
\end{itemize}
\section{More Implementation Details}
All experiments are implemented on Detectron2~\cite{wu2019detectron2}, we employ Mask2Former~\cite{Cheng_2022_mask2former} with ResNet-50 as our baseline.
Due to limited GPU memory, the backbone pre-trained by Vision-Language Models~\cite{radford2021_CLIP, li2023CLIP_Surgery} is frozen.
For optimization, we utilize AdamW~\cite{loshchilov2017decoupled} with a learning rate of 0.0001 and a weight decay of 0.05.
Following~\cite{kim2023TLDR, ding2023hgformer, bi2024CMFormer}, we apply data augmentation techniques, including random flipping, color jittering and random cropping with a size of 512$\times$1024.
During testing, images are resized to a minimum size of 1024 and a maximum of 2048.
A total of 19 semantic categories are used for training and validation.
\section{More Quantitative Analysis}
\textbf{Strategy of TDST module.} In Tab.~\ref{tab:abl_tdst}, we conducted an ablation study to evaluate the impact of different style transformation strategies on performance.
Firstly, directly fusing the original image features with the weighted style difference features yielded an average mIoU of 47.03.
By applying a controllable constraint using the hyperbolic tangent function (Tanh) to the style difference features, the average mIoU improved by +1.11.
Secondly, the low-frequency amplitude spectrum of the image features, obtained after Fourier transformation, contains the overall style and global features of the image.
However, its direct use in weighted fusion yield poor results.
This is due to the sensitivity of the low-frequency amplitude spectrum to noise from extreme values in the style difference features.
By incorporating Tanh controllable constraints, the average mIoU achieved the best results (\textit{i.e.}, 49.11).
\\
\textbf{Complexity of Models.} 
In Tab.~\ref{tab:complexity}, we compare the complexity of our method with existing domain generalization semantic segmentation (DGSS) methods. We use trainable parameters (Train Param), floating-point operations per second (FLOPs), and inference time as metrics.
Compared to Prior methods, our method significantly reduces the number of trainable parameters by freezing the backbone while remaining competitive in computational complexity and inference time.
\begin{table}[t]
\centering
\begin{adjustbox}{max width=0.47\textwidth}
\begin{tabular}{cc|ccccc}
    \toprule
    & & \multicolumn{5}{c}{Trained on GTAV (G)} \\
    \midrule
    Feature & Tanh & $\rightarrow$ C & $\rightarrow$ B & $\rightarrow$ M & $\rightarrow$ S & Avg.\\
    \midrule
    \multirow{2}{*}{Original} & $\times$ & 49.22 & 43.01 & 54.71 & 41.18 & 47.03\\
    & $\checkmark$ & 50.35 & 43.77 & 56.04 & 42.39 & 48.14\\
    \multirow{2}{*}{Low-freq.} & $\times$ & 50.41 & 43.13 & 56.25 & 42.16 & 47.99\\
    & $\checkmark$ & \textbf{51.72} & \textbf{44.67} & \textbf{56.98} & \textbf{43.08} & \textbf{49.11}\\
    \bottomrule
\end{tabular}
\end{adjustbox}
\caption{Ablation studies on Text-Driven Style Transform (TDST). "Feature" indicates whether the style transformation utilizes the original image features (Original) or the low-frequency components of the image features (Low-freq.) after Fourier transformation.}
\label{tab:abl_tdst}
\end{table}
\begin{table}[!htp]
\centering
\begin{adjustbox}{max width=0.47\textwidth}
\begin{tabular}{c|ccc}
    \toprule
    Method & Train Param$\downarrow$ & FLOPs (G)$\downarrow$ & Times (ms)$\downarrow$ \\
    \midrule
    DPCL~\cite{yang2023DPCL} & 56.46M & 1188.64 & 721.89\\
    HGFormer~\cite{ding2023hgformer} & 48.19M & 524.30 & 218.20 \\
    CMFormer (Bi et al.~\citeyear{bi2024CMFormer}) & 48.30M & 537.30 & 156.57\\
    \midrule
    SCSD (Ours) & 22.09M & 648.70 & 151.86 \\
    \bottomrule
\end{tabular}
\end{adjustbox}
\caption{Computational overhead comparison of different methods. Testing is conducted using images of size 2048×1024 on a single NVIDIA RTX 3090 GPU. We averaged the inference time over 500 trials. ``Train Param'' means trainable parameters.}
\label{tab:complexity}
\end{table}
\section{More Qualitative Analyses}
\begin{figure*}[!htp]
\centering
\begin{subfigure}[b]{0.24\textwidth}
    \centering
    \includegraphics[width=\textwidth]{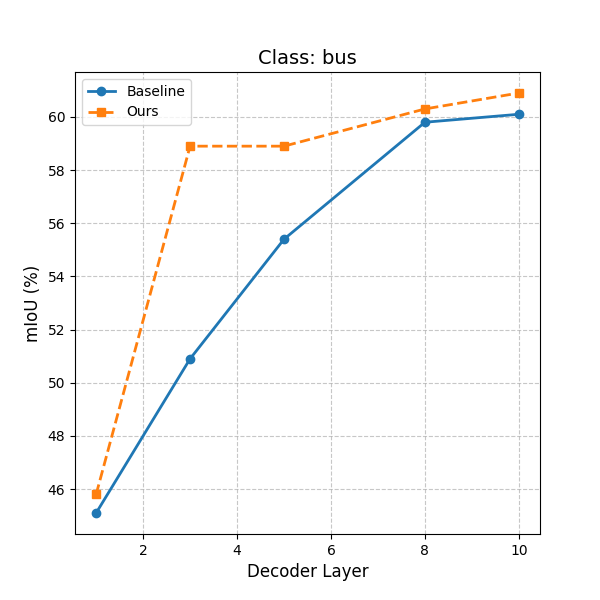}
    \caption{bus}
    \label{fig:bus}
\end{subfigure}
\begin{subfigure}[b]{0.24\textwidth}
    \centering
    \includegraphics[width=\textwidth]{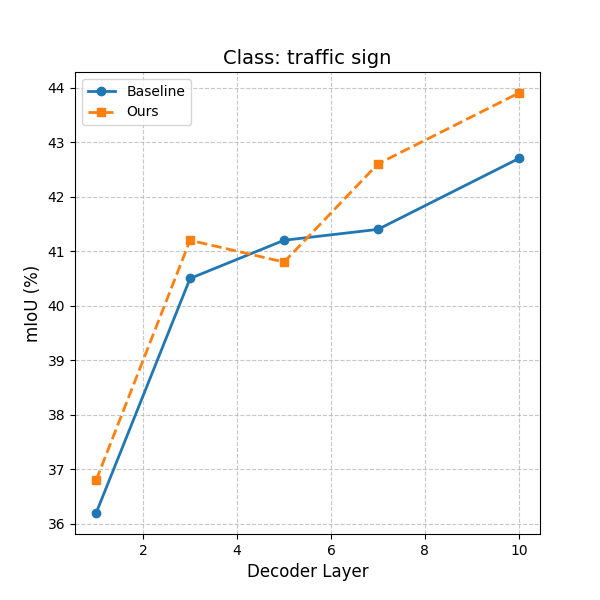}
    \caption{traffic sign}
    \label{fig:traffic_sign}
\end{subfigure}
\begin{subfigure}[b]{0.24\textwidth}
    \centering
    \includegraphics[width=\textwidth]{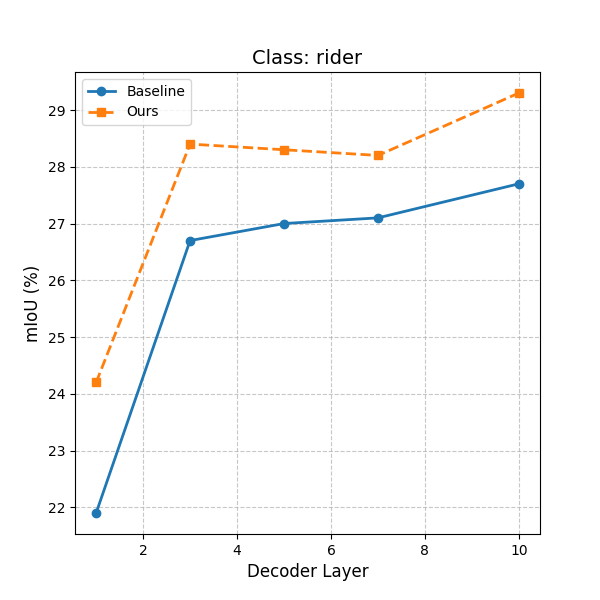}
    \caption{rider}
    \label{fig:rider}
\end{subfigure}
\begin{subfigure}[b]{0.24\textwidth}
    \centering
    \includegraphics[width=\textwidth]{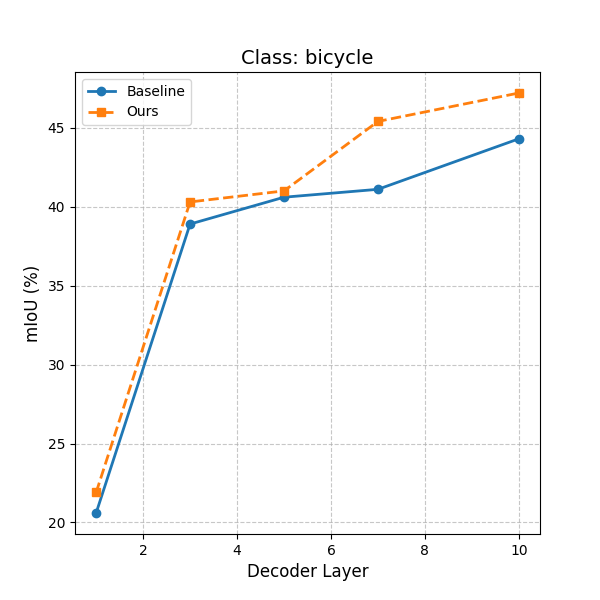}
    \caption{bicycle}
    \label{fig:bicycle}
\end{subfigure}
\caption{Comparison of the prediction results between our method (with SQB) and the baseline (without SQB) across four rare categories (\textit{i.e.}, bus, traffic sign, rider and bicycle) at different decoder layers under the G$\rightarrow$C setting.}
\label{fig:SQB_four_class}
\end{figure*}

\begin{figure*}[!htp]
\centering
\includegraphics[width=\textwidth]{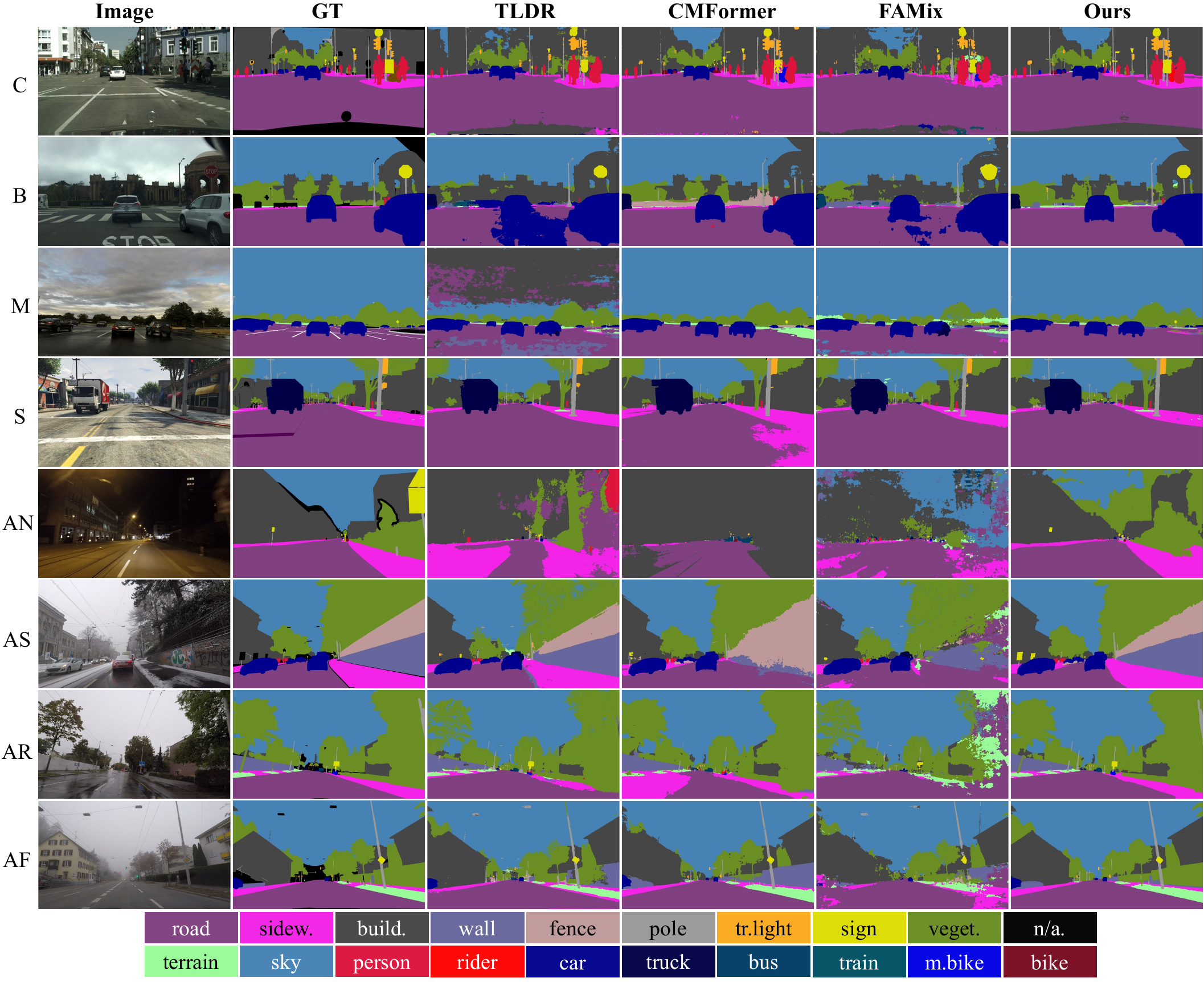}
\caption{Unseen domain segmentation prediction of existing domain generalized semantic segmentation methods~\cite{kim2023TLDR, bi2024CMFormer, fahes2024FAMix} and our proposed SCSD under the G$\rightarrow$\{C, B, M, S\} setting and adverse condition setting (\textit{e.g.}, night (AN), snow (AS), rain (AR) and fog (AF)).}
\label{mask_pred}
\end{figure*}

\textbf{Robustness of Senmantic Query Booster.}
In Fig.~\ref{fig:SQB_four_class}, we compare the prediction results of the baseline without SQB and our SCSD across different decoder layers for four rare categories (\textit{i.e.}, bus, traffic sign, rider and bicycle).
We use mIoU as the evaluation metric.
The observations reveal several key insights: first, our method equipped with SQB consistently outperforms the baseline in overall performance.
Secondly, as shown in Fig.~\ref{fig:bus}, the improvement for the ``bus'' category is particularly noticeable at the third decoder layer.
This is attributed to the object queries learning the semantic associations between images and text, thereby facilitating accurate cross-domain semantic consistency prediction.
Lastly, as the number of decoder layers increases, the mIoU gradually approaches saturation, yet it consistently remains higher than the baseline.
\\
\textbf{Segmentation results.}
In Fig.~\ref{mask_pred}, we present the semantic segmentation results on four unseen domains (\textit{i.e.}, Cityscapes~\cite{cordts2016cityscapes}, BDD-100K~\cite{yu2020bdd100k}, Mapillary~\cite{neuhold2017mapillary}, and SYNTHIA~\cite{ros2016synthia}) and one unseen adverse condition domain (\textit{i.e.}, ACDC~\cite{sakaridis2021acdc}).
To demonstrate the effectiveness of our proposed method, we also provide the results of TLDR~\cite{kim2023TLDR}, CMFormer~\cite{bi2024CMFormer}, and FAMix~\cite{fahes2024FAMix} for comparison.
In Fig.~\ref{mask_pred} (Column 3 and 5), TLDR and FAMix exhibit suboptimal performance on certain unseen domains, leading to segmentation ambiguity.
In Fig.~\ref{mask_pred} (Column 4), CMFormer performs poorly in adverse condition domains, particularly under night conditions, where it fails to segment the ``sky'' accurately.
In contrast, our proposed SCSD accurately segments and correctly classifies instances, even under adverse conditions.
These results highlight the strong generalization capability of SCSD.

\end{document}